# Sparse Depth-Guided Attention for Accurate Depth Completion: A Stereo-Assisted Monitored Distillation Approach

Jia-Wei Guo[1], Hung-Chyun Chou[1], Sen-Hua Zhu[2], Chang-Zheng Zhang[2], Ming Ouyang[2], Ning Ding[3]

*Abstract*—This paper proposes a novel method for depth completion, which leverages multi-view improved monitored distillation to generate more precise depth maps. Our approach builds upon the state-of-the-art ensemble distillation method, in which we introduce a stereo-based model as a teacher model to improve the accuracy of the student model for depth completion. By minimizing the reconstruction error of a target image during ensemble distillation, we can avoid learning inherent error modes of completion-based teachers. We introduce an Attention-based Sparse-to-Dense (AS2D) module at the front layer of the student model to enhance its ability to extract global features from sparse depth. To provide self-supervised information, we also employ multi-view depth consistency and multi-scale minimum reprojection. These techniques utilize existing structural constraints to yield supervised signals for student model training, without requiring costly ground truth depth information. Our extensive experimental evaluation demonstrates that our proposed method significantly improves the accuracy of the baseline monitored distillation method.

## I. INTRODUCTION

Scene depth perception has found numerous applications in fields such as robotics, automatic driving, and augmented reality [1, 2, 3]. Various sensors, such as LiDAR, stereo cameras, and RGB-D, can provide scene depth information [1, 4, 5]. However, these sensors have several limitations, including high costs, limited measurement distance, and high-power consumption. In recent years, there has been a growing focus on learning-based depth estimation, which has led to the development of many efficient and accurate technologies [4, 6, 7]. One of the most popular research directions in deep learning is depth completion, which aims to estimate dense pixel-wise depth based on an extremely sparse map [8, 9].

Deep learning-based methods for depth completion have shown remarkable performance and led to a development trend in recent years. Prior works have used convolutional networks [10] or simple auto-encoders [11] trained on certain batches of images and corresponding sparse depths to complete missing depth information. A traditional method in this category is to exploit dual encoders to obtain detailed features of an RGB image as well as its corresponding sparse depth map, and then fuse them using a decoder [12, 13]. Recent research on depth completion tends to use more complex network architectures and learning strategies. In addition to feature fusion from images and sparse depth, researchers have tried to introduce surface normal [14], affinity matrix [15], residual depth maps [16], and other techniques into their methodologies. To deal with the problem of lacking supervised pixels, some researchers have focused on knowledge distillation [17], multi-view geometric constraints [18, 19], and adversarial regularization [20]. For example, in [21], a blind ensemble distillation of teacher models was used to yield a distilled depth result with minimum reconstruction error and a confidence map, which can be used to train more accurate student models by selectively learning the best pixel-wise depth estimation from the ensemble and avoiding learning the teachers' error modes.

The student can only rely on unsupervised losses to obtain the correct model parameters when the procedure of assembling teachers' results yields high photometric reprojection errors. To resolve this problem, a novel stereo-based teacher is introduced to provide supervised signals for the student's training. An AS2D module is then introduced in the student model to enhance its ability to acquire global features from sparse depth. Additionally, the method leverages multi-view depth consistency between time-sequential frames to further enhance the student's model convergence. The training process is improved by incorporating multi-scale minimum reprojection among decoder layers. Experimental evaluations on the KITTI benchmark dataset show the effectiveness of our proposed method, which is better than state-of-the-art depth completion methods across four metrics. These metrics are Mean Absolute Error (MAE), Root Mean Square Error (RMSE), Mean Absolute Error of the inverse depth (iMAE) and Root Mean Square Error of the inverse depth (iRMSE).

## II. RELATED WORKS

### A. Learning-Based Stereo Estimation

Stereo estimation algorithms commonly rely on computing the similarity between corresponding pixels in a calibrated stereo image pair to infer depth information. This procedure enables a 1-D search to estimate the disparity or scaled inverse depth between the two images. Recently, learning-based methods have become popular for stereo estimation, which typically involves four key modules: computing matching cost, aggregating the cost, optimizing result, and refining final disparity.

*This work was supported by National Key R&D Program of China No. U2013202 and the Guangdong Basic and Applied Basic Research Foundation under Grant No. 2022A1515011139. Corresponding author: Hung-Chyun Chou, Email: zhouhongjun@cuhk.edu.cn.

Authors 1 are with Special Robot Center, Shenzhen Institute of Artificial Intelligence and Robotics for Society, 14-15F, Tower G2, Xinghe World, Rd Yabao, Longgang District, Shenzhen, Guangdong, 518129, China.

Authors 2 are with Huawei Cloud Computing Technology Co., Ltd.

Author 3 is with Shenzhen Institute of Artificial Intelligence and Robotics for Society, and Institute of Robotics and Intelligent Manufacturing, The Chinese University of Hong Kong, Shenzhen, Shenzhen, Guangdong, 518172, China.

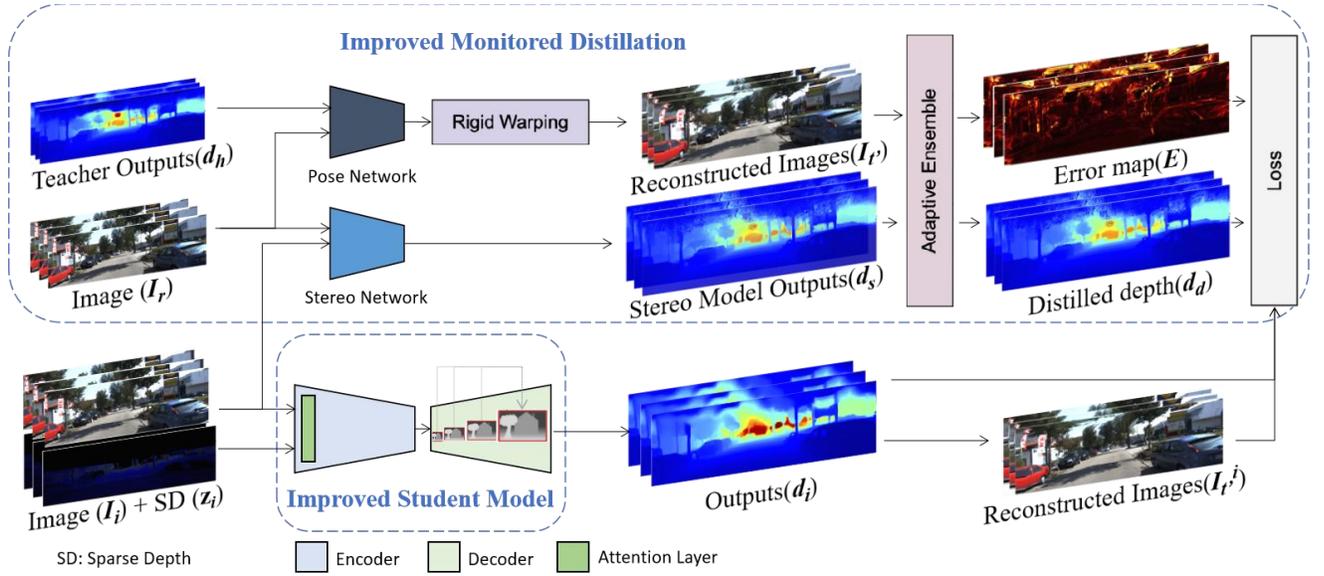

Figure 1. The framework of improved monitored distillation.

Traditionally, stereo estimation studies have worked on matching cost computation using convolution layers and then refining the disparity result using semi-global matching (SGM) [22]. However, in the recent, end-to-end network have become popular that can generate whole disparity results without the need for an explicit feature-matching module or post-processing [23, 24]. For example, PSM-Net [25] proposed an end-to-end deep learning framework for its stereo matching that does not require post-processing. It introduced a pyramid pooling module into the image features, which further incorporates global context information to improve feature extraction accuracy. In our work, we utilize a stereo-based learning model to enhance the accuracy of supervised information during monitored knowledge distillation.

### B. Supervised/Unsupervised Depth Completion

The challenge of depth completion is to accurately densify the sparse depth in the early layers of the model. To address this challenge, [12] designed a Sparse-to-Dense (S2D) module that extracts near and far features from sparse depth using different kernel sizes. This module generates a dense representation of the sparse depth. Based on this module, we design an AS2D module to further capture global features from sparse depth to densify the sparse depth.

Typical supervised depth completion methods require ground truth information as supervised signals for model training, which can be difficult to obtain. On the other hand, unsupervised depth completion methods use inherent structure cues related to the pose between adjacent images to improve model accuracy during training. Most of these methods focus on minimizing photometric errors between input images and their corresponding synthetic images reconstructed from different views, or the gap between the estimation and sparse depth, by designing suitable losses [12, 18, 19]. Other researchers have exploited synthetic data to gain priors on the shapes representing a scene [26, 27].

However, the accuracy of structure information is crucial for achieving better depth completion accuracy during training. To address this, we propose using multi-view depth consistency to supervise the model training, which not only enhances the depth completion accuracy but also improves the accuracy of structure information during training.

### C. Knowledge Distillation

Knowledge distillation is a method that utilizes a simpler student model to learn from a larger, more complicated teacher model under teacher's supervision. This method has been applied to depth estimation with effective results in recent research [28, 29]. For example, [1] used cycle inconsistency and knowledge distillation to conduct unsupervised monocular depth estimation, which the student network is a part of the teacher network. Additionally, [21] leveraged ensemble learning to adaptively select the best pixel-wise depth result from each teacher model and avoid learning error modes from these models.

In this work, we draw inspiration from [18, 21] and exploit multi-scale minimum reprojection to further enhance the accuracy of monitored distillation during depth completion training without requiring additional inputs.

### III. PROPOSED METHOD

#### A. Overview

This work focuses on the problem of obtaining a dense scene depth map from a single image and its corresponding sparse depth map. To solve this problem, we present a depth completion function that takes as input an RGB image $I_t$, its sparse depth map $z_t$, and the corresponding camera intrinsic $K_t$, and yields a dense depth map $d_t$. The function aims to imitate the implicit mapping between the input and output data by leveraging the inherent structure cues present in the scene.

$$d_t := f_\theta(I_t, z_t, K_t) \in \Re^{H \times W} \qquad (1)$$

During the training phase, we assume access to a set of synchronized image-depth pairs $\{I_i, z_i\}$, where $I_i$ corresponds to a set of temporally adjacent images $[I_{t-1}, I_t, I_{t+1}]$, and $z_i$ denotes the corresponding sparse depth map for each image. In addition, we access the corresponding image $I_r$ that is registered with image $I_t$, which enables us to use stereo-based depth estimation techniques. Furthermore, we assume that a set of N teacher depth outputs $\{d_h^i\}$, $i \in N$ related to image $I_t$ is available from publicly pre-trained models. During training, we input synchronized batches of images and sparse depth maps into the student model, which outputs a set of dense depth maps $\{d_i\}$, where $i$ represents the temporal index of the input image, i.e., $i \in [t-1, t, t+1]$. The student model employs the depth completion function described in (1) to generate the dense depth maps. Meanwhile, we generate reconstructed images $I_t^j$ corresponding to $I_t$ by applying an image warping function.

$$I_t^j = f_\omega(I_t, d_h^i) \qquad (2)$$

The difference of photometric reprojection between $I_t$ and $I_t^j$ is utilized to generate a set of error maps $E^i$. Then, these error maps can be used to validate the correctness of teacher outputs $\{d_h^i\}$. By selecting the best pixel-wise values with the highest confidence from $\{d_h^i\}$ adaptively, we can construct the distilled depth $d_d$ from these error maps. After that, we update the distilled depth by comparing the confidence of depth values with stereo outputs $d_s$ from stereo-based depth model estimation.

We integrate an AS2D module into the front layers of the student model, which enhances its ability to extract global features from the sparse depth map. During the training phase, we incorporate multi-view depth consistency by utilizing the depths from adjacent frames ($d_{t-1}$, $d_t$, $d_{t+1}$) to supervise the student model. This approach helps to enhance both the precision of depth completion and structure information simultaneously. Furthermore, we introduce the multi-scale minimum reprojection to enhance the monitored distillation for depth completion accuracy without requiring any additional inputs.

### B. Stereo Constraint Supervision

In the monitored distillation module, we leverage depth completion models as teachers for depth distillation, as proposed in [21]. To further enhance the accuracy of distilled depth in knowledge distillation, we incorporate a stereo-based end-to-end depth estimation model [25]. By inputting a pair of the left and right RGB images, this model generates a dense depth map, $d_s$, by minimizing an energy function in (3), leading to the derivation of a disparity map, $D$.

$$E(D) = \Sigma\, C(x, d_x) \qquad (3)$$

Here, $C(x, d_x)$ represents the matching cost of pixel $x = (i, j)$ in a left image and its corresponding pixel $y = (i, j - d_x)$ from a right image, where $d_x = D(x)$ represents the disparity value at pixel $x$. Stereo-based methods, by utilizing distinctive feature extraction and fusion techniques with abundant texture information from both left and right images, frequently yield depth values that are more precise compared to depth completion methods, especially for scene details. Figure 2 shows that the depth output from the stereo-based method is smoother than that from the depth completion method. Therefore, we leverage the stereo outputs $d_s$ to update the distilled depth $d_d$ through comparing the confidence of depth values.

$$d_d^i = d_s^i \text{ if } E_d^i < E_s^i \qquad (4)$$

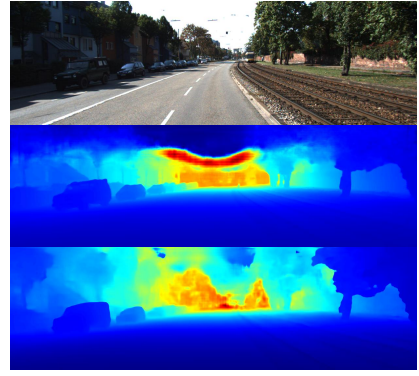

Figure 2. Depth outputs from depth completion method (middle) and stereo-based method (bottom).

### C. Attention Based Sparse-to-Dense Module

Expanding on the work presented in [12], we integrate an attention mechanism [30] (illustrated in Fig. 3) following the 1 × 1 convolution pooling operation. This integration enhances the model's ability to capture global features from the sparse map input. Firstly, we employ two sets of min pooling layers and max pooling layers to independently extract near and far features using different kernel sizes. These features are then concatenated and further processed through three 1 × 1 convolutions to learn the trade-offs between preserving detailed features and densifying the adjacent area. Subsequently, channel and spatial attention layer are used to learn more global sparse depth features for depth completion. Finally, a 3 × 3 convolution is employed to produce quasi-dense depth, which can be fed into the next layer to improve the precision of the final dense depth.

Following [12], we use five different small-size kernels in the min pooling layers to extract near features and two different big-size kernels in the max pooling layers to extract far features. After 1 × 1 convolution processing, the intermediate output $\mathbf{F}$ is further processed with a channel attention function $\mathbf{A_c} \in \Re^{C \times 1 \times 1}$ in (5) and a spatial attention function $\mathbf{A_s} \in \Re^{1 \times H \times W}$ in (6).

$$\mathbf{F}^\wedge = \mathbf{A_c}(\mathbf{F}) \cdot \mathbf{F} \qquad (5)$$

$$\mathbf{F}^\vee = \mathbf{A_s}(\mathbf{F}^\wedge) \cdot \mathbf{F}^\wedge \qquad (6)$$

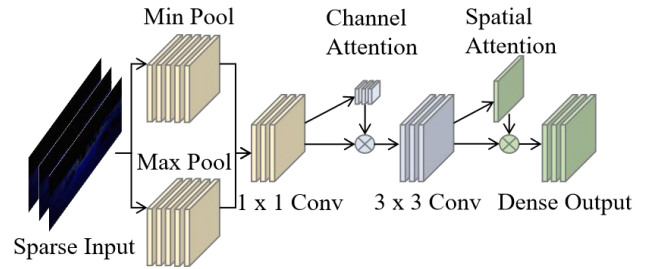

Figure 3. Attention Based Sparse-to-Dense Module.

### D. Multi-view Depth Consistency

Existing depth estimation methods do not consider the depth consistency between all adjacent views. In our proposed approach, we aim to incorporate this information by inputting both the current image and it's all adjacent images into the student model during training. This results in the generation of the corresponding depth outputs $\{d_i\}$, $i \in [t\text{-}1, t, t\text{+}1]$. These depth outputs are then used to reconstruct their corresponding adjacent views through the warping function. To further improve the accuracy of our approach, we construct photometric and structure loss functions that calculate the errors between all reconstructed adjacent images.

$$L_{photometric} = \Sigma(|I_x - I_{x|y}|) \quad (7)$$

$$L_{structure} = \Sigma(1 - \psi(I_x, I_{x|y})) \quad (8)$$

Here, $x, y \in [t\text{-}1, t, t\text{+}1]$, $x \neq y$ and $\psi$ represents structural similarity index distance, SSIM. Due to the fact that student network and pose network are responsible for the errors, the errors can be supervision information to supervise model training for better accuracy of depth completion. This is particularly useful when the monitored teacher distillation results in undesirable depth output with low confidence values [21]. Furthermore, the photometric and structure loss functions used for calculating these errors can also improve the accuracy of structure information during training.

*E. Multi-scale Minimum Reprojection*

Inspired by techniques for improving depth quality [18], we incorporate multi-scale reprojection supervision into the loss function during depth model training. The approach involves upsampling the low-resolution depth outputs, which are the intermediate outputs, to match the highest resolution input image. Based on (2), (7) and (8), we then gain the photometric and structure errors between the upsampled depth outputs and the highest resolution view as depicted in Fig. 5. Similar to patch matching, this method involves low-resolution disparity values being responsible for directly reconstructing specific blocks of pixels in the highest resolution input image. By adopting this approach, each scale resolution layer is guided towards achieving precise reconstruction of the highest resolution target image, thereby enhancing the overall quality of the depth output.

To reduce the influence of occluded pixels, which may cause high photometric error in multi-view feature matching, we introduce the minimum operation into the photometric and structure losses over all adjacent images. To obtain the final loss functions, denoted by (9) and (10) respectively, this operation is applied to the per-pixel photometric and structure loss, as defined in (7) and (8). The minimum operation selects the smallest loss value among the adjacent views for each pixel, which helps to avoid the negative influence of occlusion on the loss calculation and thus improves the accuracy of the depth estimation.

$$L_{photometric} = \Sigma(\min_{per\text{-}pixel}|I_x - I_{x|y}|). \quad (9)$$

$$L_{structure} = \Sigma(1 - \min_{per\text{-}pixel}(\psi(I_x, I_{x|y}))). \quad (10)$$

## IV. EXPERIMENTS

To assess the performance of our proposed method, we conduct an analysis using publicly available benchmarks and compare it with previously published methods. Our experiments are conducted without utilizing ground truth information during the training phase.

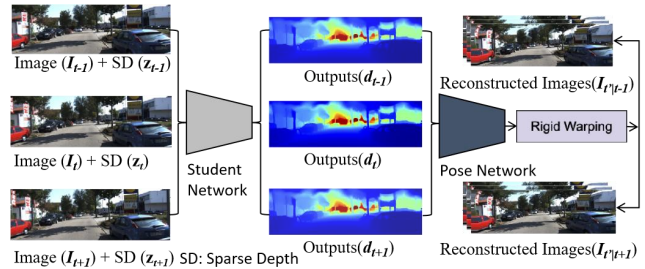

Figure 4. Multi-view depth consistency.

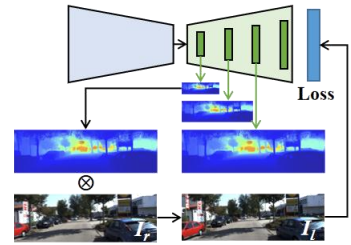

Figure 5. Multi-scale photometric reprojection.

*A. Experimental Setup and Benchmark*

We conduct experiments on the large-outdoor KITTI driving datasets and the indoor VOID datasets. KITTI comprises approximately 86,000 raw 1242x375 image frames and their corresponding synchronized sparse depth maps. These images were captured using two video cameras and a LiDAR sensor, with the valid area of sparse depth covering only about 5% of the image space. To validate our approach, we use 1,000 designated samples and their corresponding ground truth depths, along with semi-dense data. And VOID contains a series of 640 × 480 images, sparse_depth maps and ground_truth maps, which cover various indoor and outdoor scenes, such as classrooms and gardens. The ground_truth maps are obtained using a VIO system, and the sparse depth maps consist of approximately 1500 depth points with an approximate density of 0.5%. We select around 45,000 groups of images and sparse depth maps for training and 800 groups for testing.

For our distillation models, we adopt the approach of [21] and use PENet [31], MSG_CHN [32], ENet [31], and NLSPN [33] as depth completion teacher ensembles on KITTI and VOID. Additionally, we introduce PSMNet [25] as a stereo-based depth estimation teacher for KITTI to further improve our distillation method. We analyze the performance of our proposed method quantitatively using several metrics commonly used in previous works [21], as presented in Table 1. Here, we define P as the total count of pixels in the test maps, and $d_e$ and $d_{gt}$ as the estimated depth and ground truth depth, respectively.

Table 1. Evaluation metrics.

| Metric | Definition |
|---|---|
| MAE | $(\Sigma|d_e - d_{gt}|) / P$ |
| RMSE | $(\Sigma|d_e - d_{gt}|^2 / P)^{1/2}$ |
| iMAE | $(\Sigma|1/d_e - 1/d_{gt}|) / P$ |
| iRMSE | $(\Sigma|1/d_e - 1/d_{gt}|^2 / P)^{1/2}$ |

## B. Baselines for Ablation

To carry out the ablation study presented in results, we compare our proposed method with the following baselines.

Stereo-based Learning (SL): A stereo-based model is introduced into the basic framework, referred to as Baseline, proposed in [21]. The stereo outputs $d_s$ from the stereo-based depth model estimation are used to optimize the distilled depth $d_d$.

AS2D: An attention mechanism is incorporated into the basic student network, which includes a channel attention function $A_c$ and a spatial attention function $A_s$.

Multi-view Depth Consistency (MC): Based on the basic framework, multi-view adjacent images and their corresponding sparse maps are input into the student model. The model outputs the corresponding depth outputs $\{d_i\}$, $i \in [t-1, t, t+1]$, which are used to construct the photometric loss.

Multi-scale Minimum Reprojection (MR): Multi-scale reprojection supervision is introduced into the loss function through depth model training with the per-pixel minimum operation.

Our Proposed Method (Ours): The proposed method combines stereo-based learning (SL), AS2D, multi-view depth consistency (MC), and multi-scale minimum reprojection (MR) in the basic framework simultaneously.

We perform a comprehensive analysis of each baseline method to figure out the influence of each component to the overall performance of the proposed method.

## C. Results

To demonstrate the effectiveness of our proposed method, we perform an ablation study on the KITTI and VOID datasets. The quantitative results obtained on the KITTI dataset are presented in Table 2.

In Table 2, the sub-module methods proposed in this work achieve at least two better metric values than the baseline method, which was trained using the KITTI dataset. Furthermore, the method of incorporating a stereo-based learning model outperforms the baseline method in all four metrics. Figure 6 presents the distribution of the teacher models that are used to construct the distilled depth. The figure shows that the stereo-based learning teacher model contributes to improving the depth accuracy of a significant percentage of pixels, by around 10%. Our proposed method, which incorporates three sub-module methods, outperforms the baseline method across all metrics and results in significant improvement in the depth maps, as demonstrated in the qualitative comparisons on KITTI.

In Table 3 and 4, we present a comparison on KITTI with several state-of-the-art methods, considering both unsupervised learning-based, including AdaFrame, SynthProj, ScaffNet and KBNet, and supervised learning-based, such as SS-S2D, DeepLiDAR, ENet, PENet and NLSPN methods. Our proposed method is better than all unsupervised learning-based methods in all four metrics. Moreover, in Table 4, our method shows great performance among supervised learning-based methods, even outperforming SS-S2D and DeepLiDAR. These results highlight the effectiveness of our proposed method for depth completion and its potential to improve upon existing state-of-the-art methods.

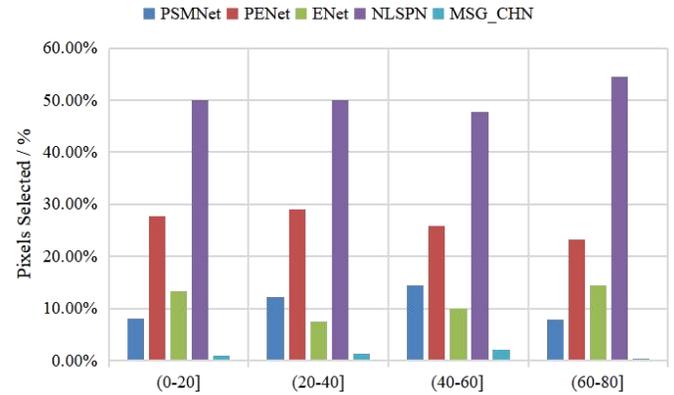

Figure 6. Teacher Model Selection Distribution.

Table 2. Evaluation metrics on KITTI.

| Method | MAE | RMSE | iMAE | iRMSE |
|---|---|---|---|---|
| Baseline | 223.021 | 808.190 | 0.953 | 2.300 |
| MR | 224.588 | 816.997 | 0.943 | 2.269 |
| MC | 222.715 | 810.013 | 0.955 | 2.286 |
| SL | 222.177 | 806.855 | 0.949 | 2.280 |
| AS2D | 219.612 | 801.689 | 0.924 | 2.211 |
| Ours | 217.881 | 800.503 | 0.920 | 2.203 |

In Figure 8, it can be observed that our proposed method yields depth maps on KITTI with improved details compared to the baseline method. Specifically, our method effectively improves the accuracy of depth estimation for various objects such as cars, poles, road lamps, and others. This suggests that our proposed method can capture more subtle details in the scene and generate more accurate depth maps.

Table 3. Unsupervised method comparisons on KITTI.

| Method | MAE | RMSE | iMAE | iRMSE |
|---|---|---|---|---|
| AdaFrame | 291.620 | 1125.670 | 1.160 | 3.320 |
| SynthProj | 280.420 | 1095.260 | 1.190 | 3.530 |
| ScaffNet | 280.760 | 1121.930 | 1.150 | 3.300 |
| KBNet | 256.760 | 1069.470 | 1.020 | 2.950 |
| Ours | 217.881 | 800.503 | 0.920 | 2.203 |

Table 4. Supervised method comparisons on KITTI.

| Method | MAE | RMSE | iMAE | iRMSE |
|---|---|---|---|---|
| SS-S2D | 249.950 | 814.730 | 1.210 | 2.800 |
| DeepLiDAR | 226.500 | 758.380 | 1.150 | 2.560 |
| Ours | 217.881 | 800.503 | 0.920 | 2.203 |
| ENet | 216.260 | 741.300 | 0.950 | 2.140 |
| PENet | 210.550 | 730.080 | 0.940 | 2.170 |
| NLSPN | 199.590 | 741.680 | 0.840 | 1.990 |

Table 5 presents comparisons of results among several methods on the VOID dataset. It can be observed that our unsupervised method, based on the AS2D module, outperforms most of the supervised methods, except for NLSPN. And our method is better than the baseline method in three metrics, indicating that the ability of our AS2D module to extract global features significantly improves the accuracy of depth completion. It can be observed from Figure 7 that our AS2D module effectively reduces depth errors in areas where there is a lack of sparse depth points.

Table 5. Comparisons on VOID.

| Method | MAE | RMSE | iMAE | iRMSE |
| --- | --- | --- | --- | --- |
| SS-S2D | 178.850 | 243.840 | 80.120 | 107.690 |
| ENet | 46.900 | 94.350 | 26.780 | 52.580 |
| PENet | 34.610 | 82.010 | 18.890 | 40.360 |
| Baseline | 29.670 | 79.780 | 14.840 | 37.880 |
| Ours (AS2D) | 29.126 | 82.152 | 13.857 | 35.666 |
| NLSPN | 26.740 | 79.120 | 12.700 | 33.880 |

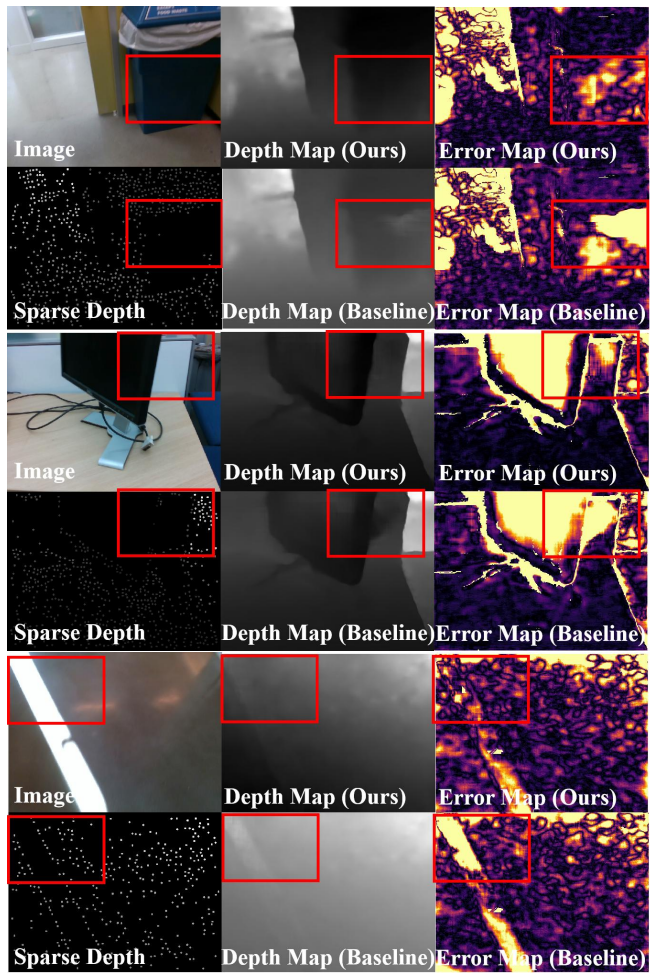

Figure 7. Qualitative comparisons on VOID (Top: the proposed method, Bottom: baseline).

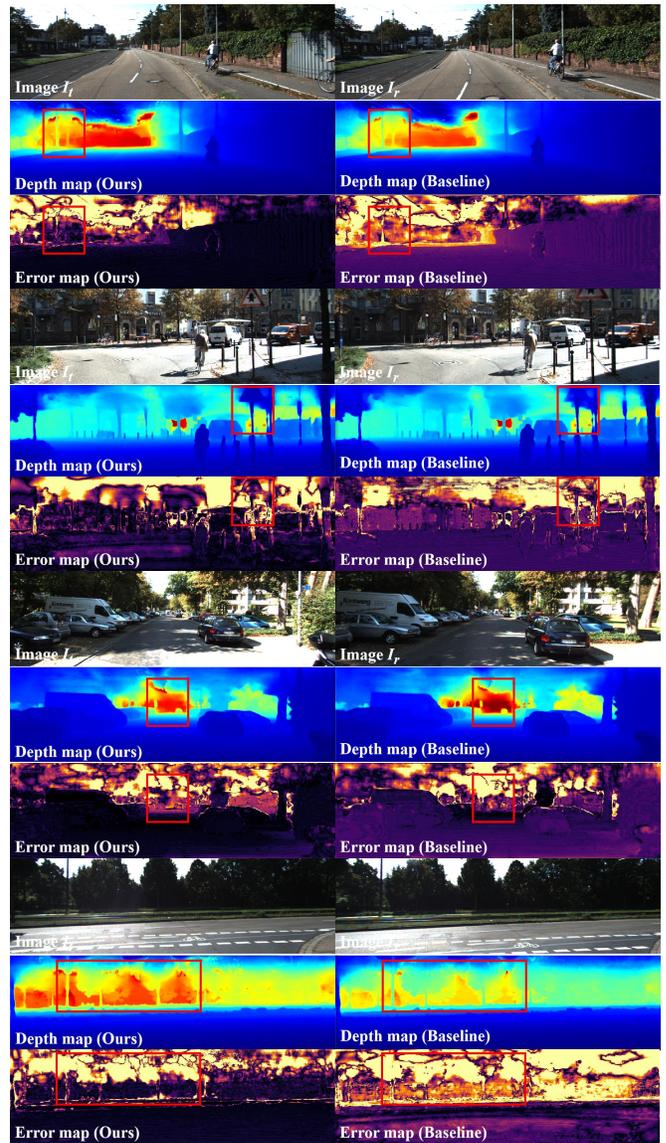

Figure 8. Qualitative comparisons on KITTI (LHS: the proposed method, RHS: baseline).

## V. CONCLUSION

The paper presents a new approach to depth completion based on multi-view improved monitored distillation, which generates more accurate depth maps. The method utilizes a publicly available stereo-based model as a teacher model to enhance the accuracy of the ensemble distillation. And an AS2D module is introduced into the student network to obtain more precise global features. Additionally, the proposed method leverages multi-view depth consistency and multi-scale minimum reprojection to provide self-supervised information, which further enhances the accuracy of the depth maps. The advantage of the proposed method is that it takes advantage of existing structure constraints to yield supervised signals for student model training without the need for ground truth information. The experimental results on KITTI demonstrate that our proposed approach effectively improves the accuracy of the baseline method of monitored distillation, with a reduction in MAE from 223.021 to 217.881, RMSE

from 808.190 to 800.503, iMAE from 0.953 to 0.920 and iRMSE from 2.300 to 2.203. However, the proposed method has limitations in handling some scenes such as high reflectivity, rapidly changing lighting conditions, and transparent objects. These limitations can be addressed in future research.